\newcommand{\ee}[1]{\mbox{e}^{#1}}

\newcommand{\dx}{\text{d}x}

\newcommand{\wdfc}{\mathcal{W}}

\documentclass[10pt,twocolumn,letterpaper]{article}

\usepackage{cvpr}
\usepackage{times}
\usepackage{epsfig}
\usepackage{graphicx}
\usepackage{amsmath}
\usepackage{amssymb}

\usepackage{subfig}
\usepackage{multirow}

\usepackage[pagebackref=true,breaklinks=true,letterpaper=true,colorlinks,bookmarks=false]{hyperref}

\cvprfinalcopy 

\def\cvprPaperID{1893} 

\begin{document}

\title{Augmenting Light Field to model Wave Optics effects}

\author{Se Baek Oh$^1$\\
$^1$Mechanical Eng. MIT\\
77 Massachusetts avenue,\\
Cambridge, MA\\
{\tt\small sboh@mit.edu}
\and
George Barbastathis$^{1,2}$\\
$^2$SMART Centre, MIT\\
3 Science Drive 2, Singapore\\
{\tt\small gbarb@@mit.edu}
\and
Ramesh Raskar\\
Media Lab. MIT\\
20 Ames street, Cambridge, MA\\
{\tt\small raskar@media.mit.edu}
}

\maketitle

{\emph{This is work in progress. Your comments and suggestions are highly welcome.}}

\begin{abstract}
The ray--based 4D light field representation cannot be directly used to analyze diffractive or phase--sensitive optical elements. In this paper, we exploit tools from wave optics and extend the light field representation via a novel ``light field transform". We introduce a key modification to the ray--based model to support the transform. We insert a ``virtual light source", with potentially negative valued radiance for certain emitted rays. We create a look--up table of light field transformers of canonical optical elements. The two key conclusions are that (i) in free space, the 4D light field completely represents wavefront propagation via rays with real (positive as well as negative) valued radiance and (ii) at occluders, a light field composed of light field transformers plus insertion of (ray--based) virtual light sources represents resultant phase and amplitude of wavefronts. For free--space propagation, we analyze different wavefronts and coherence possibilities. For occluders, we show that the light field transform is simply based on a convolution followed by a multiplication operation. This formulation brings powerful concepts from wave optics to computer vision and graphics. We show applications in cubic--phase plate imaging and holographic displays.
\end{abstract}


\section{Introduction\label{sec:intro}}
The light field (LF) is a four--variable parameterization of the plenoptic function~\cite{LevoyLF96,GortlerLF96} describing the radiance of a ray propagating along $\theta_x$ and $\theta_y$ directions at $x$ and $y$. The ray--based 4D LF representation is based on simple 3D geometric principles and has led to a range of new algorithms and applications. They include digital refocusing, depth estimation, synthetic aperture, and glare reduction. However, the LF representation is inadequate to describe interactions with diffractive or phase--sensitive optical elements (\ie, phase masks or holography). In such cases, Fourier optics principles are often used to represent wavefronts with additional phase information. It is known that the Wigner distribution function (WDF) is a counterpart of the LF in wave optics. The WDF describes the local spatial frequency spectrum of light, where the local spatial frequency $u$ corresponds to the ray angle $\theta$. 
\begin{figure}
\centering
\includegraphics[width=0.8\linewidth]{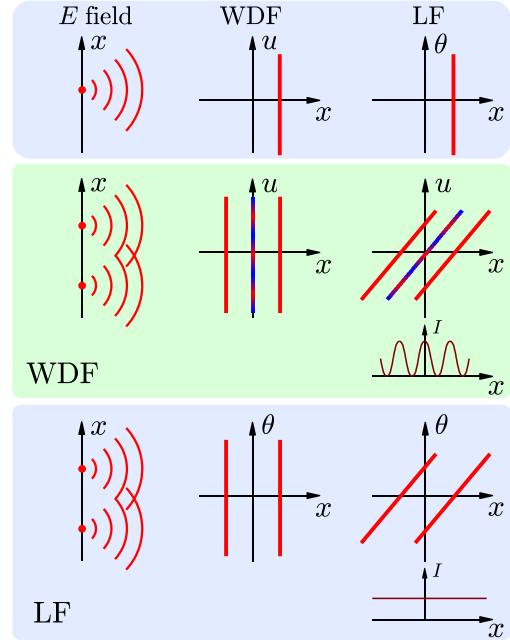}
\caption{\emph{(Top)} Wavefront, WDF, and LF representation of a point source. \emph{(Middle)} The WDF representation and \emph{(Bottom)} The LF representation of wavefront at and after propagation in Young's experiment. In the WDF representation, we see that an additional oscillatory term is introduced at the midpoint of the two point sources, which produces interference after propagation.}\label{fig:summary}
\end{figure}
Figure~\ref{fig:summary} represents an important observation. In some cases, \eg, for a point object, the WDF and the LF exhibit identical properties. However, in other cases, \eg, in Young's experiment, the WDF and LF differ at occluders as well as after finite propagation.

\begin{figure}
\centering
\includegraphics[width=0.9\linewidth]{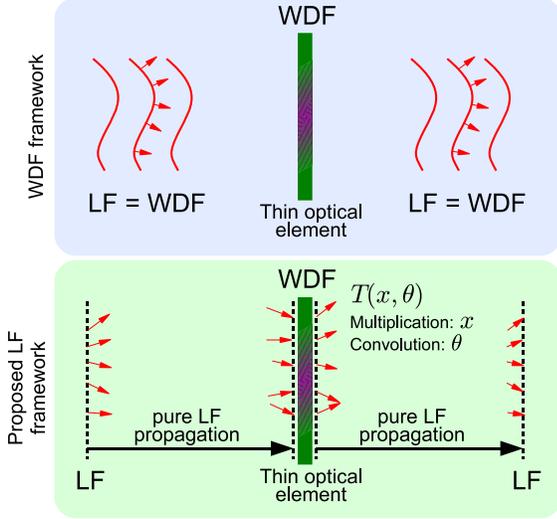}
\caption{The augmented LF transformer concept. (Top) In free space, we make WDF and LF equivalent (via negative radiance values). (Bottom) At thin optical element, we show that a LF transformer performs multiplication in the spatial domain and convolution in the angular domain.}\label{fig:summary2}
\end{figure}

In this paper, we exploit relevant tools in wave optics and present an augmented LF framework to handle diffraction and phase--sensitive imaging. Figure~\ref{fig:summary2} summarizes the key idea. We develop a simple toolbox to explain LF propagation through such materials. We introduce a notion of virtual light sources, for which the radiance of a ray is real (positive as well as negative) valued. We show that the augmented LF representation is sufficient to describe general wavefront propagation.

For simplicity, we explain the light propagation in flat land (\ie, in the plane of the paper). In the flat land the LF and WDF are 2D functions. The same analysis applies to the real 3D world, where the LF and WDF are 4D functions.

\subsection{Contribution}
For a more comprehensive LF analysis, we adapt useful tools from wave optics and present an augmented LF propagation framework. Specific technical contributions are as follows:
\begin{itemize}
\item We derive new LF propagation (for free--space) \& transformation (for occluders) equations that are as powerful as traditional wave--optics techniques.
\item We observe the following two facts: i) In free--space propagation, the WDF and LF are equivalent for any coherence state. ii) For interaction with thin elements in the optical path, transforms of incident LF plus virtual light sources produce an exact solution for the resultant LF.
\item We show applications of the formulation in devices for which LF analysis has not been used before, such as interferometers, phase plates, and holographic displays.
\end{itemize}
We hope to inspire researchers comfortable with ray--based analysis to start exploiting more complex optical elements. They can use well--understood LF concepts plus the new ray--based tools we have introduced, without worrying about complex Fourier--domain calculus common in wave--optics.
\begin{figure}
\centering
\includegraphics[width=0.9\linewidth]{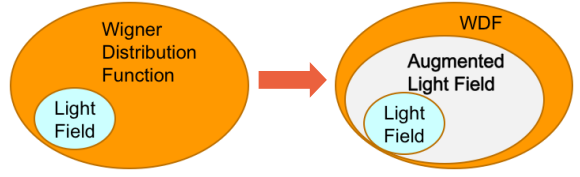}
\caption{Comparison of the LF, ALF, and WDF. The LF formulation lacks phase properties and its radiance is always positive, whereas the ALF supports partially coherent light and diffractions by introducing the LF transformers and virtual light sources. \label{fig:LF_ALF_WDF_comp}}
\end{figure}

\subsection{Scope and limitation}
Since we intend to model diffraction and phase--related optical phenomena, coherence of light should be properly treated. Conceptually, coherence indicates the ability of making interference. Coherent light, \eg,  a wavefront from a laser, has deterministic phase relations over all wavefronts, whereas incoherent light has completely random phase relations. Partially coherent state refers to any coherence state in between completely coherent and incoherent ones. The spatial frequency spectrum of incoherent light is extended to the evanescent cut--off in principle; the propagation direction is not well defined or equivalently incoherent light propagates along all directions. Hence, incoherent light may be regarded to be superposition of infinite numbers of plane waves with random phase delays. 
Although the WDF can be defined for partially coherent light, our formulations and representations mostly deal with coherent light, because our formulations for coherent light is easy to understand and can be extended into partially coherent/incoherent light in straightforward fashion. Hence, in this paper we briefly mention the significance of coherence state and explain the effect on the proposed formulations. More rigorous description of coherence can be found in~\cite{Goodman_Statistical}. 

Throughout this paper, we consider linearly polarized monochromatic light in the paraxial regime for simplicity. Introducing additional dimension for the wavelength easily extends our formulations into polychromatic light. The paraxial approximation, a common assumption in wave--optics theory as well as most practical cases, simplifies mathematical descriptions, especially the LF transformer in Sec.~\ref{sec:lightfieldtransformer}. For more rigorous polarization analysis, matrix or tensor methods can be employed as in~\cite{WDF_Alonso_NonparaxialPolarized}. We also consider the free--space propagation in homogenous medium and neglect any non--linear optical effect. 

As the virtual light sources and LF transformers are still based on ray representations, our model would have the same limitations as other ray--based models; if there is a caustic or singularity in systems, our model would not provide accurate results. 

\subsection{Related Work}

\noindent\textbf{Light fields and shield fields:} Light fields were proposed by Levoy and Hanrahan~\cite{LevoyLF96} and Gortler \etal~\cite{GortlerLF96} to characterize the propagation of rays. Several camera platforms have been developed for capturing light fields: Ives~\cite{Ives28} and Lippman~\cite{Lipmann1908} used an array of pinholes. Wilburn \etal used camera arrays~\cite{Wilburn05} and Georgiev \etal~\cite{TodorEGSR06} put both prisms and lenses in front of a camera. Adelson and Wang~\cite{Adelson_PleopticsCamera} and Ng \etal~\cite{Lightfield_RenNg_PlenopticCamera} devised plenoptic cameras consisting of a single lens and a lenslet array. Instead of the lenslet array, a sinusoidal attenuating mask was used in the heterodyne camera~\cite{Veeraraghavan07}. Beside the light field capture systems, research about light transport, cast shadows, and light field in frequency domain has been studied by Chai \etal~\cite{ChaiPlenopticSampling00}, Isaksen \etal~\cite{IsaksenReparameterizedLF00}, and Durand \etal~\cite{Fredo_LightTransport}. Shield fields were introduced to analyze attenuation of rays through occluders~\cite{Lanman_Shielfields}.\\

\noindent\textbf{Wigner distribution function:} The WDF was originally proposed by Wigner in quantum mechanics~\cite{WDF_wigner}. In optics, the WDF of light (electric--field) contains both the space and local spatial frequency information. The WDF has been exploited in analysis and design of  various optical systems: 3D display~\cite{WDF_Javidi_3Ddisplay}, digital holography~\cite{WDF_Javidi_Occluded,WDF_Stern_Sampling}, generalized sampling problems~\cite{WDF_Stern_Sampling}, and superresolution~\cite{WDF_Zalevsky_superresolution}. The WDF can also be defined for partially coherent light~\cite{WDF_Bastiaans_PartiallyCohernet} and thin optical elements such as a lens, phase mask, aperture, or grating~\cite{WDF_Bastiaans_1stOrder,WDF_Bastiaans_CircularAperture,WDF_Casteneda_PhaseGrating}. The ambiguity function, the Fourier transform pair of the WDF, corresponds to the LF in the frequency domain and has been used in understanding wavefront coding systems~\cite{CPM_Cathey_Paradigm,CPM_ExtendedDOF_Dowski95}. More details of the WDF can be found in Ref.~\cite{WDF_bastiaans_chapter}. 

In the optics community, many researchers have tried to connect radiometry and wave optics. One notable concept is the generalized radiance suggested by Walther~\cite{RadiometryCoherence_Walther}. Wolf investigated extensively and summarized its physical meaning and limitation in~\cite{CoherenceRadiometry_Wolf1978}. In computer vision and graphics communities, Zhang and Levoy recently reviewed this connection~\cite{WDF_Levoy_WDF_LF}. After the WDF was revealed as one kind of phase--space distribution functions, many different phase--space distribution functions have been proposed; angle--impact Wigner function~\cite{WDF_AngleImpactWDF} and Choi--Williams distribution function~\cite{WDF_ChoiWilliams} can potentially be employed in developing new systems and algorithms in computer vision and graphics.\\

\noindent\textbf{Ray based model for diffraction:} many different theories have been proposed to model diffraction in the context of ray--optics. The geometrical theory of diffraction (GTD) is widely known~\cite{GTD_Keller_JOSA1962}. To model diffraction at edge, the GTD exploits various laws of diffraction and computes diffraction coefficients. Since the augmented LF is utilized the WDF based on wave--optics, diffraction is automatically taken into account. More importantly, the augmented LF is implemented in the LF framework. Hence, the augmented LF is much more convenient than the GTD and provides greater versatility to researchers in computer graphics and vision communities.

\section{Augmented light field propagation framework}
\begin{figure}
\centering
\includegraphics[width=0.9\linewidth]{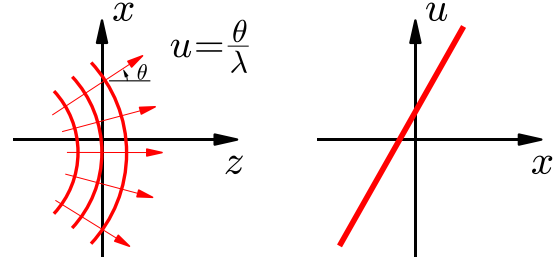}
\caption{Visualization of a wavefront and its Wigner distribution function. Rays are normal to the wavefront and the phase of the wavefront is equivalent to the local spatial frequency in the Wigner representation.}\label{fig:wave_ray}
\end{figure}

The LF is the radiance of a ray parameterized with a position $x$ and angle $\theta$. In wave optics, light is often described by the amplitude and phase of electric fields. The wavefront is defined as a surface of a constant phase in the electric fields. The wavefront and its WDF are shown in Fig.~\ref{fig:wave_ray}. \emph{Rays are always normal to the wavefront} and \emph{the phase of the wavefront is encoded in the local spatial frequency $u$.} The propagation angle $\theta$ of a ray and the local spatial frequency $u$ is related by $u = \theta/\lambda$~\cite{Goodman_Fourier}, where $\lambda$ denotes the wavelength. Hence, in free--space without any light interference, the LF representation is complete and contains the phase information of the wavefront.

The Wigner distribution function for an input $g(x)$ is defined as
\begin{equation}
\wdfc (x,u) = \int g(x+\tfrac{x'}{2})g^*(x-\tfrac{x'}{2})\ee{-i2\pi ux'}\dx',
\label{eq:wdf_def}
\end{equation}
where $g(x)$ can be either electric fields or transmittance of an optical element. Projecting the WDF along the $u$--axis yields the intensity just as in the LF. 

\begin{figure}
\centering
\includegraphics[width=0.9\linewidth]{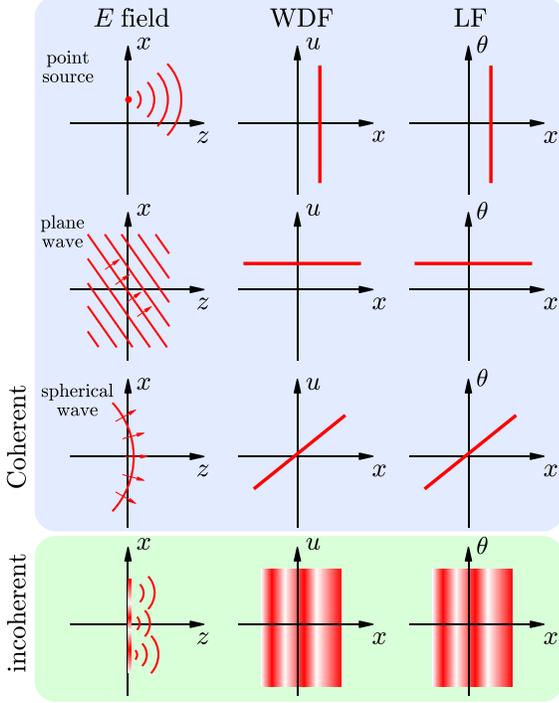}
\caption{Comparisons of both WDF and LF for few simple lights. The LF and the WDF exhibit identical fashions.}\label{fig:WDF_LF_comp}
\end{figure}
Figure~\ref{fig:WDF_LF_comp} shows wavefronts, the WDF, and the LF for a point source, plane wave, spherical wave, and incoherent light. The WDF and LF exhibit identical representations for these lights.  Note that top three shown in Fig.~\ref{fig:WDF_LF_comp} are coherent lights and the fourth one represents incoherent light of lateral radiance variation. 

\subsection{Limits of Light Field Analysis}
\begin{figure}
\centering
\includegraphics[width=0.9\linewidth]{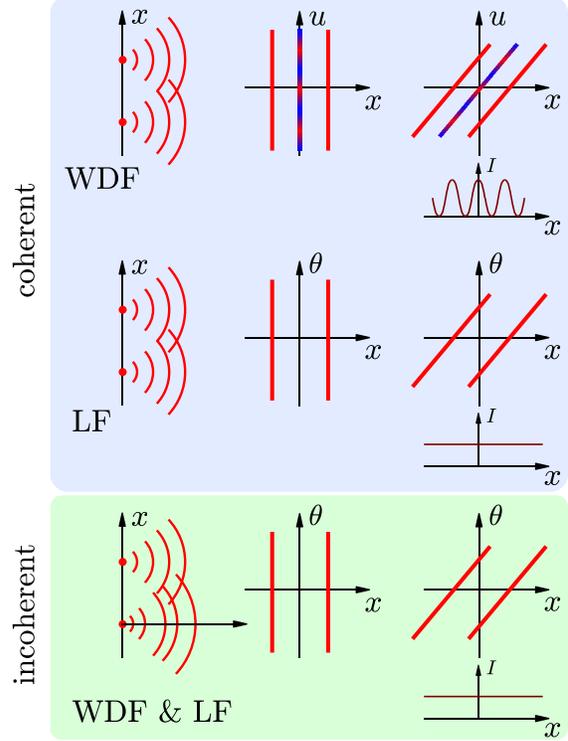}
\caption{The WDF and LF representations of Young's experiment, where the third light in the WDF produces interference. If the light is incoherent, even in the WDF, the third light diminishes and both representations predict no interference. (Color code in the WDF; red: positive, blue: negative)}
 \label{fig:Young_exp}
\end{figure}

The LF is limited for elements showing diffraction or phase sensitive behavior (\ie, phase gratings or holography). To exemplify, Young's experiment (two pinholes illuminated by a laser) is analyzed by both the WDF and LF as shown in Fig.~\ref{fig:Young_exp}.
In the WDF representations, the two point sources (from the pinholes) produce three components: two at the two pinholes' locations and the other at the middle of the two pinholes. The last component is called an interference term and obtained by mathematical manipulation of the WDF~\cite{WDF_RCastaneda_PSParCoh}. For infinitesimally small pinholes, the transmittance of the two pinholes is given by
\begin{equation}
g(x) = \delta(x-a)+\delta(x-b),
\end{equation}
where $a$ and $b$ denote the locations of the pinholes. Then, $g\left(x+\tfrac{x'}{2}\right)g^*\left(x-\tfrac{x'}{2}\right)$ is expanded as
\begin{multline}
\left[\delta\left(x-a+\tfrac{x'}{2}\right)+\delta\left(x-b+\tfrac{x'}{2}\right)\right]\\
\times\left[\delta\left(x-a-\tfrac{x'}{2}\right)+\delta\left(x-b-\tfrac{x'}{2}\right)\right]\\
=\delta(x-a)\delta(x')+\delta(x-b)\delta(x')\qquad\qquad\qquad\\
+\delta\left(x-q\right)\delta\left(x'-p\right)+\delta\left(x-q\right)\delta\left(x'+p\right),
\label{eq:twodeltas}
\end{multline}
where two new variables are defined as $p=a-b$ and $q=(a+b)/2$. Taking the Fourier transform with respect to $x'$ of eq.~\eqref{eq:twodeltas} computes the WDF of the two pinholes as
\begin{multline}
\mathcal{W}(x,u) = \delta(x-a)+\delta(x-b)\\+2\delta\left(x-\tfrac{a+b}{2}\right)
\cos\left(2\pi\left[a-b\right]u\right).
\label{eq:wdf_twopinholes}
\end{multline}
As shown in eq.~\eqref{eq:wdf_twopinholes}, the last cosine term oscillates between positive and negative values along the $u$--axis, thus it does not contribute to the intensity. In the LF description, only two light fields exist. As the light propagates, both the WDF and LF are sheared along the $x$--direction. The intensity of the fringe is computed by projection along either $u$ or $\theta$; no interference is produced in the LF.

If the light is incoherent, then both the WDF and LF representations do not have the interference term. As described earlier, the incoherent light is considered to be an infinite number of plane waves propagating along all directions with random phase delays. If the two pinholes are probed by the incoherent light, then the three components of eq.~\eqref{eq:wdf_twopinholes} are replicated an infinite number of times with all possible offsets in the $u$--axis; the interference terms are averaged out. This will be clearly understood with the incoherent light and the LF transformer of the two pinholes described in Sec.~\ref{sec:lightfieldtransformer}.

\subsection{Virtual light sources{\label{sec:virtual_src}}}
\begin{figure}
\centering
\includegraphics[width=0.9\linewidth]{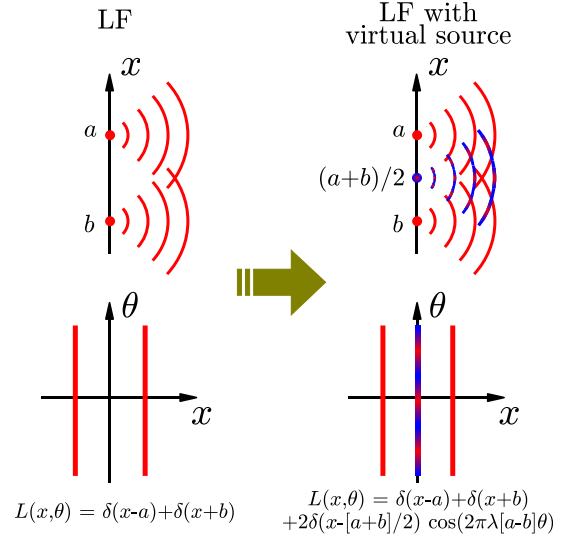}
\caption{Concept of virtual light sources for coherent light. In the LF representation, no interference is predicted. By including the virtual light sources, the LF propagation still can be used.}
\label{fig:vir_src}
\end{figure}

To rigorously use the LF description for diffraction and interference, the interference term should be included. Here, we expand the LF framework by introducing the concept of virtual light sources, which may have negative radiance.  For the case of the two--pinholes at $a$ and $b$ as shown in Fig.~\ref{fig:vir_src}, the location of the virtual light source is at $(a+b)/2$ and its radiance is $2\cos\left(2\pi[a-b]\frac{\theta}{\lambda}\right)$ along the $\theta$--axis. Since the intensity is obtained by integrating the augmented LF along the $\theta$--axis, the virtual light sources do not affect the intensity at the pinholes plane, which agrees with physical observation and intuition. Once the virtual sources are included in the LF, then the LF propagation still can be used and interference can be properly modeled by the augmented LF.
Note that, as the definition eq.~\eqref{eq:wdf_def} implies, computing the WDF of an optical element is indeed locating the virtual light sources for all the possible combinations of two points on the element.

\subsection{Light field transformer\label{sec:lightfieldtransformer}}
Next we model the LF propagation through an optical element with a LF transformer.
\begin{figure}
\centering
\includegraphics[width=1\linewidth]{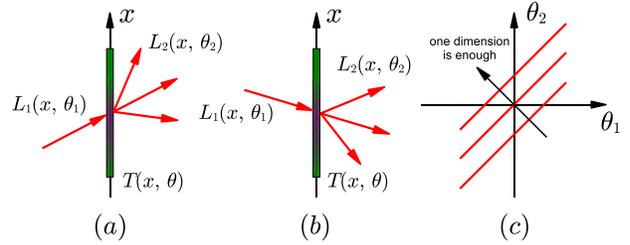}
\caption{Angle shift invariance in a thin transparency. In (a) and (b), the output rays rotate in the same fashion as the input ray rotates, which allows that one $\theta$ argument is sufficient in the LF transformer.}
\label{fig:trans_geo}
\end{figure}
As shown in Fig.~\ref{fig:trans_geo}, in the LF transformer model, a thin optical element is probed by an input LF $L_1(x_1, \theta_1)$, and an output LF $L_2(x_2, \theta_2)$ is generated. In the most generalized situation, the relation of the input and output LFs is constructed as
\begin{equation}
L_2(x_2, \theta_2) = \iint T(x_2, \theta_2, x_1, \theta_1) L_1(x_1, \theta_1) \dx_1\text{d}\theta_1,
\label{eq:8D}
\end{equation}
where $T(\cdot)$ denotes the LF transformer of the optical element. Equation~\eqref{eq:8D} indicates that the optical element introduces a 4D transform (8D in the real world) from $(x_1, \theta_1)$ to $(x_2, \theta_2)$ space. 

For a thin optical element, $x_2 = x_1$. Then, eq.~\eqref{eq:8D} becomes
\begin{equation}
L_2(x, \theta_2) = \int T(x, \theta_2, \theta_1) L_1(x, \theta_1) \text{d}\theta_1,
\label{eq:6D}
\end{equation}
where a 3D transform (6D in the real world) is involved. In most thin optical elements, they exhibit angle shift invariance in the paraxial region; \eg let us consider that an optical element produces three rays from an incoming ray of an incident angle $\theta_1$ at $x_1$ (Fig.~\ref{fig:trans_geo}(a)). As the input ray rotates, the three output rays also rotate in the same manner as shown in Fig.~\ref{fig:trans_geo}(b). Hence, the LF transformer is sufficiently described by only one $\theta$ argument and eq.~\eqref{eq:6D} is further simplified as
\begin{equation}
L_2(x, \theta_2) = \int T(x, \theta_1-\theta_2) L_1(x, \theta_1) \text{d}\theta_1.
\label{eq:4D}
\end{equation}
The LF transformer is indeed a 2D transform in the flat land (4D in the real world) as shown in Fig.~\ref{fig:trans_geo}(c). Equation~\eqref{eq:4D} is particularly interesting because it involves a multiplication along the $x$--axis but a convolution along the $\theta$--axis. Note that the LF transformer can be computed by the WDF and we present the LF transformer of canonical elements in Sec.~\ref{sec:amp_masks} and Sec.~\ref{sec:phase_masks}.

When a ray passes through an optical element, if it does not bend and only the radiance is attenuated, and then eq.~\eqref{eq:4D} becomes even simpler as
\begin{equation}
L_2(x, \theta) = S(x, \theta) L_1(x, \theta),
\label{eq:2D}
\end{equation}
where $S(x,\theta)$ is the shield field, describing attenuation by occluders~\cite{Lanman_Shielfields}.\\

\noindent\textbf{Claim 1:} At a thin interface, $L_2(x,\theta)$ is expressed by a special operation on $L_1(x,\theta)$ and a 4D LF transformer $T(x,\theta)$, in which the operation is a convolution along the $\theta$--axis and a multiplication along the $x$--axis.

\section{Propagation in Free Space{\label{sec:prop}}}
In wave optics, the free--space propagation is described by the Fresnel diffraction~\cite{Goodman_Fourier}. Applying the WDF to the Fresnel diffraction formula, we obtain the free--space propagation relation in the WDF framework, which is the $x$--shear transform~\cite{WDF_bastiaans_chapter} just as the LF propagation.\\

\noindent\textbf{Claim 2:} For the free--space propagation, the WDF and LF exhibit the identical $x$--shear transform. \\

In the case of the far--zone diffraction (Fraunhofer diffraction), the free--space propagation becomes the Fourier transform, in which the LF rotates $90^\circ$. The fractional Fourier transform~\cite{FractionalFT1_Ozaktas,FractionalFT2_Ozaktas} describes any propagation in between the far and near--zone more rigorously, where the WDF and LF both rotate by the amount of the propagation. For all coherence states, the free--space propagation is identically illustrated as the $x$--shear transform. 
\begin{figure}
\centering
\includegraphics[width=0.9\linewidth]{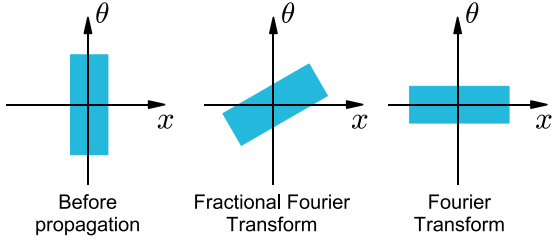}
\caption{Two more representations of free--space light propagation}
\label{fig:prop}
\end{figure}

\section{Propagation via masks}
For a thin optical element of an amplitude mask $t(x)$ and a phase mask $\exp\left\{i\phi(x)\right\}$, the LF transformer is computed by the WDF; applying eq.~\eqref{eq:wdf_def} to $g(x)=t(x)\exp\left\{i\phi(x)\right\}$. For convenience, we pre--computed the LF transformers of commonly used amplitude and phase masks in following sections.  As described in Sec.~\ref{sec:lightfieldtransformer}, an incident LF interacts with a LF transformer, presented in Table 1 and 2, and we can predict an outgoing LF by eq.~\eqref{eq:4D}. The virtual light sources are automatically introduced in this transform. For example, for two pinholes, the third term oscillating at $x=(a+b)/2$ is the virtual light source. 

\subsection{Propagation via Amplitude Masks\label{sec:amp_masks}}
\begin{figure}
\centering
\includegraphics[width=0.9\linewidth]{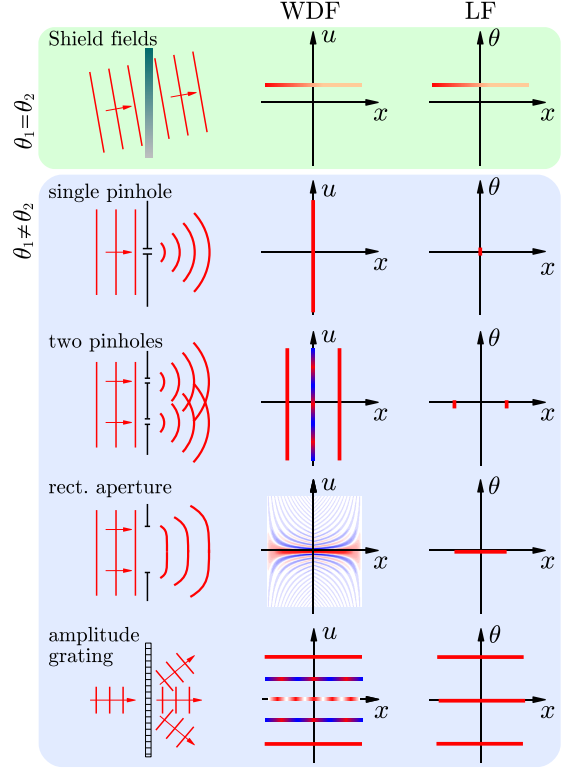}
\caption{Comparison of the WDF and LF of lights for various amplitude masks. The WDF columns represent both the LF transformers of the masks and the output LF with virtual light sources.}\label{fig:ampmask}
\end{figure}

Figure~\ref{fig:ampmask} shows some of commonly used amplitude masks, where the WDF column represents the WDF (or LF transformer) of the elements and the LF column represents the traditional LFs. Interestingly, the LF of a thin optical element is identical to the output LF when the optical element is probed by a plane wave propagating normal to the $x$--axis, because the LF of the plane wave is $\delta(\theta)$; a convolution along the $\theta$--axis produces the LF transformer itself. 

\begin{table*}
\centering
\begin{minipage}{0.7\textwidth}
\begin{tabular}{c|c}
  \hline
  amplitude mask $t(x)$& $T(x, \theta)$\\ \hline
  one pinhole &\multirow{2}{*}{$\delta(x-x_0)$}\\
  $\delta(x-x_0)$& \\ \hline
  two pinholes & \multirow{2}{*}{$\delta(x-a)+\delta(x-b)+2\delta(x-[a+b]/2)\cos\left(\tfrac{2\pi}{\lambda}(a-b)\theta\right)$}\\ $\delta(x-a)+\delta(x-b)$ &  \\ \hline
  rectangular aperture\footnote{$\Lambda$: triangle function defined in~\cite{Goodman_Fourier}; if $|x|\leq1$, $\Lambda(x)=1-|x|$, otherwise $\Lambda(x)=0$.} &  \multirow{2}{*}{
$2A\Lambda\left(\frac{x}{A/2}\right)\text{sinc}\left([2A-4|x|]\frac{\theta}{\lambda}\right)$}
  \\
  $\text{rect}\left(\frac{x}{A}\right)$ & \\ \hline
  amplitude grating & $\tfrac{1}{4}\Big[\left\{1+\frac{m^2}{2}\cos\left(\tfrac{2\pi}{p}2x\right)\right\}\delta(\theta)+$ \\
  \multirow{2}{*}{$\tfrac{1}{2}\left(1+m\cos\left(\frac{2\pi}{p}x\right)\right)$}  & $m\cos\left(\tfrac{2\pi}{p}x\right)\left\{
  \delta\left(\theta-\tfrac{\lambda}{2p}\right)+\delta\left(\theta+\tfrac{\lambda}{2p}\right)\right\}$\\
  & $+\tfrac{m^2}{4}\left\{\delta\left(\theta-\tfrac{\lambda}{p}\right)+\delta\left(\theta+\tfrac{\lambda}{p}\right)\right\}\Big]$
  \\ \hline
  coded aperture & \multirow{2}{*}{
 $\int t\left(x+\tfrac{x'}{2}\right)t^*\left(x-\tfrac{x'}{2}\right)\ee{i\frac{2\pi}{\lambda} x'\theta}\dx'$}\\
  $t(x)$  & \\
  \hline
  \end{tabular}
  \end{minipage}
  \caption{LF transformer of amplitude masks}
\end{table*}

\subsection{Propagation via Phase Masks\label{sec:phase_masks}}
\begin{figure}
\centering
\includegraphics[width=0.9\linewidth]{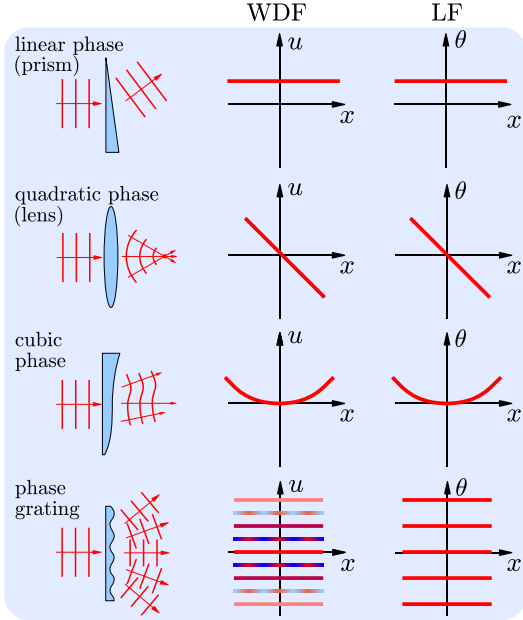}
\caption{Comparison of the WDF and LF of lights for various phase masks. The light field transformer can be easily computed from the transparency function $A(x)\ee{\phi(x)}$.}
\label{fig:phasemask}
\end{figure}

We also compute the LF transformers of various phase masks and show in Fig.~\ref{fig:phasemask}.
For optical elements with slowly varying phase variations whose complex transmittance is defined as $\exp\left\{i\phi(x)\right\}$, the LF transformer is given by~\cite{WDF_Brenner_1984}
\begin{equation}
T(x,\theta) = \delta\left(\theta-\frac{\lambda}{2\pi}\frac{\partial \phi}{\partial x}\right).
\end{equation}
This explains why the LF transformer of a cubic phase mask is a quadratic curve in sec.~\ref{sec:phase_masks}.

\begin{table*}
\centering
\begin{minipage}{0.6\textwidth}
\begin{tabular}{c|c}
  \hline
  phase mask $e^{\phi(x)}$& $T(x, \theta)$\\ \hline
  linear phase (prism) & \multirow{2}{*}{$\delta(\theta+\lambda\alpha)$}\\ $\alpha  x$ & \\ \hline
  quadratic phase (lens) & \multirow{2}{*}{$\delta\left(\theta+\frac{2\pi}{f}x\right)$}\\
  $\frac{\pi}{\lambda}\frac{x^2}{f}$ & \\\hline
  cubic phase & \multirow{2}{*}{$\delta\left(\theta-\frac{\lambda}{2\pi}\alpha x^2\right)$}\\
  $\alpha x^3$ &  \\\hline
  phase grating\footnote{$J_q$: Bessel function of the first kind, order $q$} &   $\sum_{m=-\infty}^{\infty}\Big\{\sum_{n=-\infty}^{\infty} J_{m+n}(\phi_0) J_n(\phi_0) $\\
   $\phi_0\sin\left(\frac{2\pi}{p}x\right)$ &
   $\delta\left(
\theta - \lambda[m+2n]/(2\Lambda)\right) \Big\}\exp\left\{j2\pi m x /\Lambda\right\}$\\ \hline
  phase plate\footnote{assuming slowly varying phase $\phi(x)$} & \multirow{2}{*}{$\delta\left(\theta-\frac{\lambda}{2\pi}\frac{\partial \phi}{\partial x}\right)$}\\
  $\phi(x)$ & \\ \hline
  \end{tabular}
\end{minipage}
  \caption{LF transformer of phase masks}
\end{table*}

\section{Results}
We demonstrate how to use the LF transformer and the augmented LF propagation for three specific systems.

\subsection{Airy pattern in a single lens imager}
\begin{figure}
\centering
\includegraphics[width=0.9\linewidth]{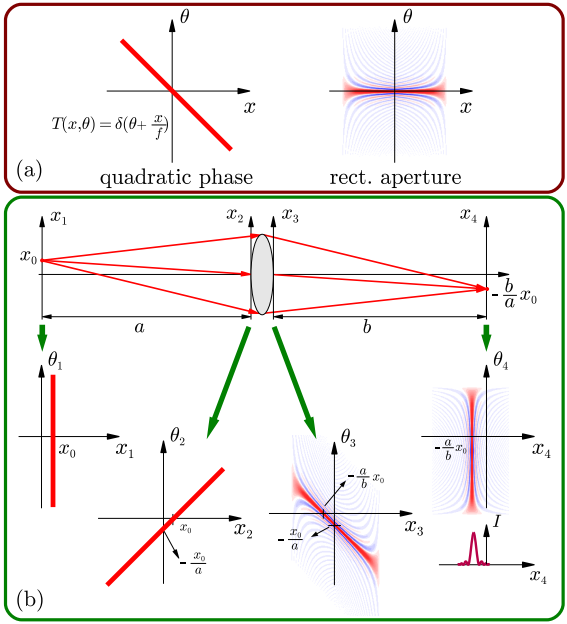}
\caption{Point spread function (Airy pattern) of a single lens imager. (\emph{a}) LF transformer of the phase and amplitude components of a lens with focal length $f$, (\emph{b}) LF shape as the propagation through the system. Note that due to the finite size aperture, the PSF is the airy pattern.}
\label{fig:single_lens_imager}
\end{figure}

For a single lens imager shown in Fig.~\ref{fig:single_lens_imager}, an ideal point spread function is an infinitesimally small point if diffraction is ignored. However, it is well known that the point spread function is indeed the Airy disk due to diffraction by the aperture. Here we explain how the augmented LF allows us modeling diffraction.

A lens, focal length $f$ and aperture size $A$,  can be decomposed of a pure phase mask of quadratic phase variation (\ie, quadratic change in optical path length as a function of $x$) and an amplitude mask of a rectangular aperture as shown in Fig.~\ref{fig:single_lens_imager}(a). Figure~\ref{fig:single_lens_imager}(b) shows how the augmented LF changes throughout the system. The LF of a point source at $x=x_0$ is $\delta(x-x_0)$ at the object plane and is sheared along the $x$--axis by the propagation to the lens. By the LF transformer shown in Fig.~\ref{fig:single_lens_imager}(a), the augmented LF transmitted the lens is a tilted blurb with some negative radiance values; the quadratic phase of the lens induces the tilt, and the finite aperture produces radiance variations. Then, the augmented LF is sheared again along the $x$--axis by the second propagation to the image plane. Integrating the augmented LF along the $\theta$--axis, we obtain the intensity of the point spread function, which is the Airy pattern in the flat land.

\subsection{Wavefront coding system (Cubic phase mask)}
We apply the same procedure to a wavefront coding system: specifically a cubic phase mask imager for extending the depth of field~\cite{CPM_Cathey_Paradigm,CPM_ExtendedDOF_Dowski95}. With the cubic phase mask, rays experience different focal lengths depending on their positions at the mask, and they bend differently compared to those in the single lens imager. The cubic phase mask reshapes the point spread function invariant to defocus; thus deconvolution in post--processing is much robust and the depth of field in processed images is extended. 

\begin{figure}
\centering
\includegraphics[width=0.9\linewidth]{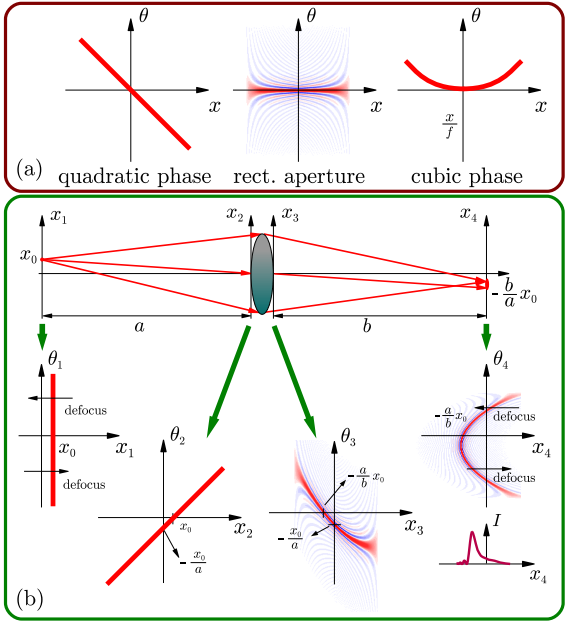}
 \caption{Point spread function of the cubic phase mask imager. (\emph{a}) LF transformer of a lens and cubic phase mask, (\emph{b}) LF shape as the propagation through the system. Note that the PSF is distorted but invariant to defocus.}
 \label{fig:CPM_imager}
\end{figure}

As shown in Fig.~\ref{fig:CPM_imager}(a), the cubic phase mask imager has three LF transformers: a lens, rectangular aperture, and cubic phase mask. The LF transformer of the cubic phase mask is a quadratic curve. Figure~\ref{fig:CPM_imager}(b) shows the system geometry and the behavior of the LF through the system.
Again we start with a point object and the LF is sheared along the $x$--axis by the propagation to the lens. By the lens and the cubic phase mask, the transformed LF becomes a curved blurb. After the second propagation, the augmented LF is transported at the image plane. Since the LF is not parallel to the $\theta$--axis, the point spread function is distorted and has asymmetric tails. 

If an object is defocused, then the LF of the object at the focus plane is not parallel to the $\theta$--axis and is rather tilted. This sheared input LF also causes the sheared output LF at the image plane in the $x$--direction; however, the intensity does not change significantly because the contributions from the upper and lower parts of the curved LF are compensated with each other. In the original derivations of the extended depth of field by using the cubic phase mask~\cite{CPM_ExtendedDOF_Dowski95,CPM_Cathey_Paradigm}, the ambiguity function was exploited to represent the OTF for various defocus simultaneously. Finally we want to mention that various phase masks have been analyzed for the extended depth of field~\cite{ExDOF_Castro,ExDOF_Sauceda,ExDOF_Sheng,ExDOF_Sherif,ExDOF_Sun,ExDOF_WDF_Sicre,WDF_Yang_EDOFanalysis,ExDOF_Bagheri}.

\subsection{Hologram Transform of Light Field}
\begin{figure}
\centering
\includegraphics[width=0.9\linewidth]{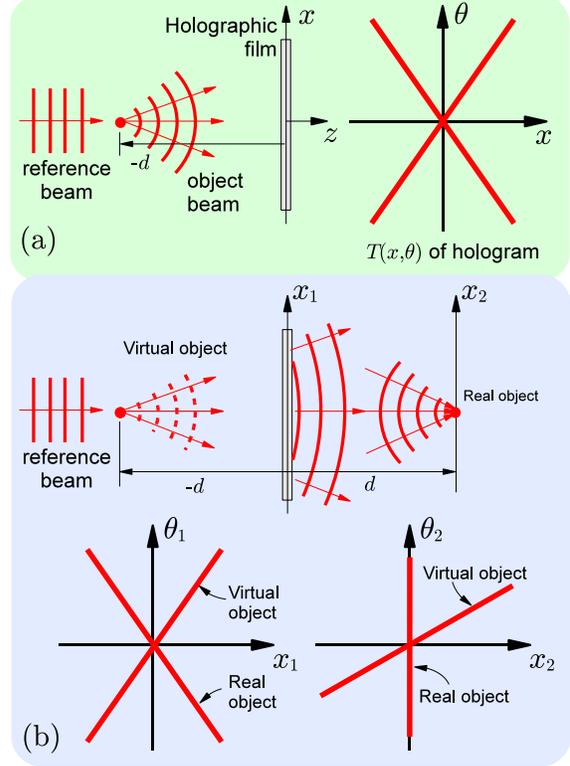}
\caption{Light field--based reconstruction from hologram of a single Lambertian scene point. \emph{(Left in (a))} Recording geometry, and \emph{(Right in (a))} the LF transformer of the recorded hologram. \emph{(Top in (b))} Reconstructing hologram with the reference wave. Two diffracted lights are produced at $z=0$ and $z=d$. \emph{(Bottom left in (b))} the LF at $z=0$, \emph{(Bottom right in (b))} the LF at $z=d$.}
\label{fig:holography}
\end{figure}

In this section, given the LF transformer of a hologram, we show how to predict an output image. For simplicity, we choose a point source as an object. Thus, a spherical wave from the point source is an object wave, and a plane wave propagating parallel to the optical axis is a reference wave as shown in Fig.~\ref{fig:holography}(a). 
To obtain the LF transformer of the hologram, we first compute the transmittance of the hologram, which is proportional to the intensity of the interference between the object and reference waves. From the geometry, the electric field of the reference and object waves in the paraxial region are given by 
\begin{eqnarray}
&E_r(x,y,z)= \exp\left\{i\frac{2\pi}{\lambda}z\right\},\quad\text{and}\qquad\qquad\label{eq:ref_wv}\\
&E_o(x,y,z)= \exp\left\{i\frac{2\pi}{\lambda}(z+d)+i\frac{\pi}{\lambda}\frac{x^2}{z+d}\right\}.\label{eq:obj_wv}
\end{eqnarray}
The intensity of the interference is proportional to
\begin{equation}
I\sim E_r^*E_o + E_r E_o^*.\label{eq:holo_sig}
\end{equation}
Here we ignored DC terms, $|E_r|^2$ and $|E_o|^2$, because they are uniform and do not contain high frequency signals. By substituting eqs.~\eqref{eq:ref_wv} and \eqref{eq:obj_wv} with eq.~\eqref{eq:holo_sig} and using $z=0$ (hologram is at $z=0$), the transmittance of the hologram is proportional to
\begin{multline}
I\sim \exp\left\{i\frac{2\pi}{\lambda}d\right\}
\exp\left\{i\frac{2\pi}{\lambda}\frac{x^2}{2d}\right\}+\\
\exp\left\{-i\frac{2\pi}{\lambda}d\right\}
\exp\left\{-i\frac{2\pi}{\lambda}\frac{x^2}{2d}\right\}.\label{eq:holo_sig2}
\end{multline}
Now we compute the WDF of eq.~\eqref{eq:holo_sig2} as
\begin{multline}
\mathcal{W}(x,u)
=\delta\left(u-\frac{x}{\lambda d}\right)+\delta\left(u+\frac{x}{\lambda d}\right)+\\
2\cos\left(\frac{2\pi}{\lambda}\left[2d+\frac{x^2}{d}-d\lambda^2u^2\right]\right),
\end{multline}
where we have not shown the last cosine term in Fig.~\ref{fig:holography} because it is a higher order oscillation term spread over the entire $x$--$\theta$ space. Finally, the LF transformer of the hologram is
\begin{multline}
T(x,\theta) = \delta\left(\theta-\frac{x}{d}\right)+\delta\left(\theta+\frac{x}{d}\right)+\\
2\cos\left(\frac{2\pi}{\lambda}\left[2d+\frac{x^2}{d}-d\theta^2\right]\right).
\end{multline}

In the reconstruction, the replica of the reference wave probes the hologram. Since the LF of the reference wave is $\delta(\theta)$, the LF of the reconstructed wave is shown in the bottom left of Fig.~\ref{fig:holography}(b). Behind the hologram, at $z>0$, two waves are produced: 1) a converging spherical wave, which is focused at $z=d$, and 2) a diverging spherical wave, which looks like originating from a point object at $z=-d$. If an observer looks at the hologram from the other side of the original point object from any distance $(z>0)$, the diverging spherical wave produces a virtual image of the original point object. If a screen is placed at $z=d$, one will also observe a real image. Note that the two components in the LF transformer of the hologram correspond to the virtual and real images as shown in the right panel of Fig.~\ref{fig:holography}(a). 

Here we simplified the holography extremely. In practice, since coherent light is used in both recording and reconstruction, the high--order oscillation terms, often called cross--terms or interference terms in the optics community, should be considered as well. More rigorous analysis of holography with the WDF can be found in~\cite{WDF_holography_Sheridan,WDF_Holography_Testorf}.

\section{Conclusion}
Geometrical optics, commonly used in computer vision, and wave optics, used in diffraction and interference analysis, employ different formulations. Yet they describe the same propagation of light in different contexts. There are many interesting complementary concepts in these two areas; \eg, the Shack--Hartmann sensor for sensing aberrations in a wavefront, is an identical concept and uses the same optics as the plenoptic cameras detect radiance along different angles. The similarity of the LF and the WDF could be one of the most fascinating connections between computer vision/graphics and optics communities. 

Whereas the LF describes many optical phenomena well, it is limited to incoherent light and inadequate to describe diffraction and phase-sensitive optical elements. In this paper, we employed wave optics to broaden the scope of the LF and the augmented LF can model diffraction and interference. For the free-space propagation, the LF and the WDF exhibit the same $x$--shear transform. To account for diffraction due to either amplitude or phase variations, we introduced the concept of the virtual light sources with varying, possibly negative, radiance along different angles. Once the virtual light sources are included properly in the augmented LF framework, one can continue to use the LF propagation in any optical systems. We also introduced the concept of the LF transformer, which describes the relation between the input and output LF for optical elements. We assumed a thin transparency and angle shift--invariance, which is true for most conventional optical elements, the LF transformer is represented by a 2D lookup table in the flat land. We pre--computed the 2D lookup tables for canonical optical elements such as an aperture, a lens, gratings, and various phase masks. Since these results already include the virtual sources, we are able to extend the LF framework under diffraction and phase--sensitive elements. The benefit is that optical systems previously considered beyond the analysis abilities of the LF, such as phase masks and holograms, can now be used in computer vision and graphics applications with simple modifications.

Although our examples are described with coherent light, incoherent or partially coherent light are much more interest; because the WDF is more powerful for incoherent and partially coherent than for coherent light~\cite{WDF_Alonso_Cho,WDF_Testorf_thesis}. Moreover, most applications in computer vision and graphics deal with partially coherent and incoherent light. As we described earlier, the augmented LF can be extended into partially coherent as well as incoherent light. 

There are several future directions of exploration. One obvious application is rendering. Since the proposed formulation can incorporate geometrical characteristics with wave property of light, it would provide more realistic rendering results for surfaces involving diffraction and interference~\cite{CGV_RenderingBiologicalIridescences,CGV_RenderingIridescentNaturalObjects,CGV_RenderingOpticalDisks}. Compared to wave optics, geometric optics assumes infinitely small wavelengths. We can create a continuum of solutions for the LF when this assumption is not valid. One can derive the LF transformer equation from the WDF by assuming infinitely small wavelengths. In the rectangular aperture shown in Fig.~\ref{fig:ampmask}, as the wavelength decreases, the WDF of the aperture is squeezed down along the $u$--axis and it eventually becomes $\delta(\theta)\text{rect}(x/A)$. We would also like to explore the LF transformers beyond thin transparencies or occluders with angle shift invariance. Other manipulations will include volumetric objects with 6D or 8D transforms that will result in 3D or 4D lookup tables for optical elements.

Finally our goal is not only to introduce WDF to the computer vision and graphics communities but also to support more rigorous analysis of augmented light field that could lead to a new class of applications. We hope this work will inspire researchers in optics as well as in computer vision/graphics to develop new tools and algorithms based on joint exploration of geometric and wave optics concepts; \eg, utilizing useful concepts and techniques from wave optics brought more realistic results and different data representation to computer graphics applications~\cite{CGV_Ziegler_BRLFHologram, CGV_Ziegler_WaveFramework}.

\section*{Acknowledgement}
We would like to thank Markus Testorf for very useful comments. We encourage readers to look at two special literatures edited by him which inspired us: \emph{``Selected papers on phase--space optics"} (SPIE's milestone series in optical science \& engineering) and \emph{``Phase--space representations in optics"} (Special issue of \emph{Applied Optics} in 2008).

{\small
\bibliographystyle{ieee}

}

\end{document}